\documentclass[runningheads]{llncs}
\usepackage{graphicx}
\usepackage{caption}
\usepackage{subcaption}
\usepackage{tikz}
\usepackage{comment}
\usepackage{amsmath,amssymb} % define this before the line numbering.
\usepackage{color}
\usepackage{multirow}
\usepackage{tabularray}
\usepackage{makecell}
\usepackage{orcidlink}

\usepackage[accsupp]{axessibility}  % Improves PDF readability for those with disabilities.

\begin{document}
% \renewcommand\thelinenumber{\color[rgb]{0.2,0.5,0.8}\normalfont\sffamily\scriptsize\arabic{linenumber}\color[rgb]{0,0,0}}
% \renewcommand\makeLineNumber {\hss\thelinenumber\ \hspace{6mm} \rlap{\hskip\textwidth\ \hspace{6.5mm}\thelinenumber}}
% \linenumbers
\pagestyle{headings}
\mainmatter
\def\ECCVSubNumber{5512}  % Insert your submission number here

\newcommand{\shortname}{BATMAN}
\newcommand{\fullname}{Bilateral Attention Transformer in Motion-Appearance Neighboring space}

\title{\shortname: Bilateral Attention Transformer in Motion-Appearance Neighboring Space for Video Object Segmentation} % Replace with your title

% INITIAL SUBMISSION 
\begin{comment}
\titlerunning{ECCV-22 submission ID \ECCVSubNumber} 
\authorrunning{ECCV-22 submission ID \ECCVSubNumber} 
\author{Anonymous ECCV submission}
\institute{Paper ID \ECCVSubNumber}
\end{comment}
%******************

% CAMERA READY SUBMISSION
%\begin{comment}
\titlerunning{Accepted as an oral paper by ECCV 2022}
% If the paper title is too long for the running head, you can set
% an abbreviated paper title here
%
\author{Ye Yu\inst{1}\orcidlink{0000-0002-4650-5894} \and
Jialin Yuan\inst{2}\orcidlink{0000-0003-1390-2233} \and
Gaurav Mittal\inst{1}\orcidlink{0000-0002-6942-3891} \and
Li Fuxin\inst{2}\orcidlink{0000-0001-7578-9622} \and
Mei Chen\inst{1}\orcidlink{0000-0003-3540-1229}
}

\authorrunning{Y. Yu et al.}
% First names are abbreviated in the running head.
% If there are more than two authors, 'et al.' is used.
%
\institute{Microsoft \\
\email{\{yu.ye,gaurav.mittal,mei.chen\}@microsoft.com}\\
%\url{http://www.springer.com/gp/computer-science/lncs} \and
\and Oregon State University\\
\email{\{yuanjial,lif\}@oregonstate.edu}}
%\end{comment}
%******************
\maketitle

\begin{abstract}
Video Object Segmentation (VOS) is fundamental to video understanding.
Transformer-based methods show significant performance improvement on semi-supervised VOS.
However, existing work faces challenges segmenting visually similar objects in close proximity of each other.
In this paper, we propose a novel {\fullname}~(\shortname) for semi-supervised VOS. 
It captures object motion in the video via a novel optical flow calibration module that fuses the segmentation mask with optical flow estimation to improve within-object optical flow smoothness and reduce noise at object boundaries. This calibrated optical flow is then employed in our novel bilateral attention, which computes the correspondence between the query and reference frames in the neighboring bilateral space considering both motion and appearance. Extensive experiments validate the effectiveness of {\shortname} architecture by outperforming all existing state-of-the-art on all four popular VOS benchmarks: Youtube-VOS 2019 (85.0\%),  Youtube-VOS  2018 (85.3\%),  DAVIS  2017Val/Test-dev (86.2\%/82.2\%), and DAVIS 2016 (92.5\%).
\footnotetext[2]{At Oregon State University, Jialin Yuan and Li Fuxin are supported in part by NSF grant 1911232.}

\end{abstract}

\section{Introduction}
Video Object Segmentation (VOS) is fundamental to video understanding with broad applications in content creation, content moderation, and autonomous driving. In this paper, we focus on the semi-supervised VOS task, where we segment target objects in each frame of the entire video sequence (query frames) given their segmentation masks in the first frame (reference frame) only. Moreover, the task is class-agnostic in that we do not have any class annotation for any object to be segmented in either training or testing phases. The key challenge in semi-supervised VOS is how to propagate the mask from the reference frame to all the query frames in the rest of the sequence without any class annotation.

Due to the absence of class-specific features, VOS models need to match features of the reference frame to that of the query frames both spatially and temporally to capture the class-agnostic correspondence and propagate the segmentation masks. Previous methods attempt to store features from preceding frames in memory networks and match the query frame through a non-local attention mechanism~\cite{STM,STCN}, or compute a global-to-global attention through an encoder-decoder transformer~\cite{TransVOS}, or propagate and calibrate features from the reference frame to the query frames using a propagation-correction scheme~\cite{RPCMVOS}. These methods employ a global attention mechanism to establish correspondence between the full reference frame and the full query frame. This can lead to failure in distinguishing the target object(s) from the background particularly when there are multiple objects with a similar visual appearance. 
A spatial local attention is proposed in \cite{AOT} to mitigate this problem, where the attention is only computed between each query token and its surrounding key tokens within a spatial local window. However, it still suffers from incorrectly segmenting visually similar objects in close proximity of each other.

In addition to spatial correspondence, it is essential to match features temporally for optimal object segmentation across video frames.
To this end, some VOS methods~\cite{RMNet,SegFlow} leverage optical flow to capture object motion. ~\cite{RMNet} warps the memory frame mask using optical flow before performing local matching between memory and query features based on the warped mask, while ~\cite{SegFlow} simultaneously trains the model for object segmentation and optical flow estimation by bidirectionally fusing feature maps from the two branches. However, these methods are not able to perform optimally as optical flow is usually noisy and warping features/masks to match objects across frames accumulates errors in both optical flow and segmentation mask along the video sequence.

\begin{figure}[t]
\centering
\includegraphics[width=\textwidth]{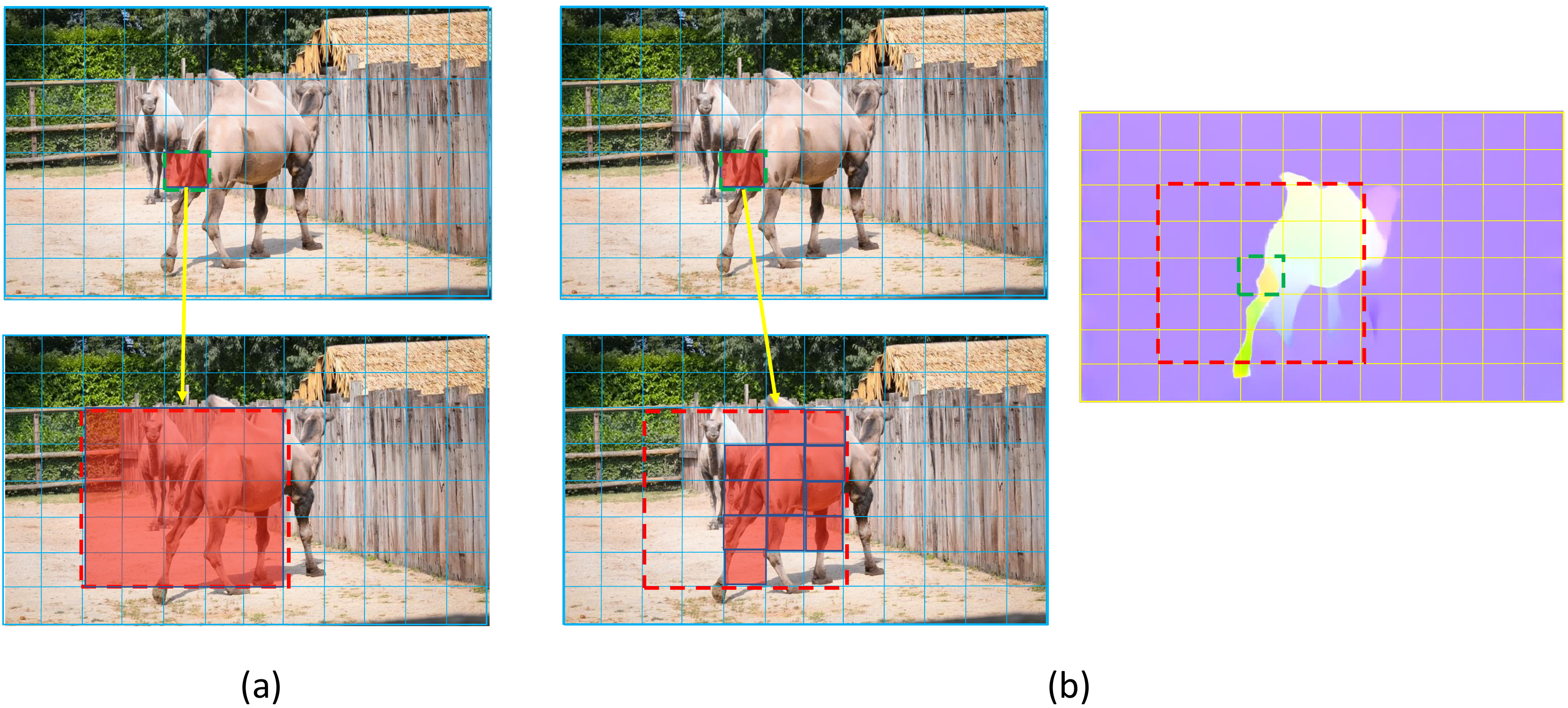}
\caption{Spatial local attention vs. bilateral attention. (a) Conventional spatial local attention. For any given token in the query frame (top), compute the attention with the neighboring tokens within a predefined fixed local window from the memory frame (bottom). (b) Our proposed bilateral attention. Given a token in the query frame (top), adaptively select the most relevant tokens (bottom), based on the distance in the bilateral space of appearance and motion (right), for cross attention computation}
\label{fig:attention}
\end{figure}

To overcome the above challenges, we propose {\fullname} (\shortname). {\shortname} introduces a novel bilateral attention module that computes the local attention map between the query frame and memory frames with both motion and appearance in consideration. Unlike the conventional spatial local attention mechanism (Fig.~\ref{fig:attention}(a)) that computes the attention within a predefined fixed local window, our bilateral attention adaptively computes the local attention based on the tokens' spatial distance, appearance similarity, and optical flow smoothness, as shown in Fig.~\ref{fig:attention}(b). Observing that optical flow may be especially noisy for fast-moving object(s),  {\shortname} introduces a novel optical flow calibration module that leverages the mask information from the memory frame to smooth the optical flow within the same object while reducing noise at the object boundary.

We conduct extensive experiments on four popular VOS benchmarks: 
Youtube-VOS 2019 \cite{xu2018youtube}, Youtube-VOS 2018 \cite{xu2018youtube}, DAVIS 2017 \cite{pont20172017}, and DAVIS 2016 \cite{perazzi2016benchmark} to validate the {\shortname} architecture. We show that {\shortname} achieves superior performance on all benchmarks and outperforms all previous state-of-the-art methods. We summarize the main contributions of our work below,

\begin{itemize}
    \item[$\bullet$] A novel bilateral attention module that computes attention between query and memory features in the bilateral space of motion and appearance, which improves the correspondence matching by adaptively focusing on relevant object features while reducing the noise from the background.
    \item[$\bullet$] A novel optical flow calibration module that fuses the object segmentation mask and the initial optical flow estimation to smooth the within-object optical flow and reduce noise at the object boundary.
    \item[$\bullet$] Incorporating the optical flow calibration and bilateral attention mechanisms, we design a novel {\shortname} architecture. {\shortname} establishes new state-of-the-art performance on Youtube-VOS 2019 / 2018 and DAVIS 2017 / 2016 benchmarks. To the best of our knowledge, {\shortname} is the first work to compute attention in the bilateral space of motion and appearance for VOS.
\end{itemize}

\section{Related work}
\subsubsection{Semi-supervised VOS.} The task aims to segment the particular object instances throughout the entire video sequence given one or more annotated frames (the first frame in general). Early DNN works \cite{caelles2017one,perazzi2017learning,xiao2018monet} fine-tune the pre-trained networks on the first frame using multiple data augmentations on the given mask at test time to adapt to specific instances. Therefore, these methods are extremely slow during inference due to excessive fine-tuning.
Later tracking-based works \cite{wang2019fast,khoreva2019lucid,chen2020state} adopt object tracking technologies to indicate the target location of objects for segmentation to improve inference time. However, these approaches are not robust to occlusion and drifting with error accumulated during the propagation. ``Tracking-by-detection" paradigm is introduced into VOS in \cite{huang2020fast} to take object segmentation as a subtask of tracking, in which the accuracy of tracking often limits the performance. 
To handle occlusion and drifting, matching-based methods \cite{chen2018blazingly,voigtlaender2019feelvos} perform feature matching to find objects that are similar to the target objects in the reference frames. 
STM \cite{STM} and its following works \cite{seong2020kernelized,RMNet} leverage an external memory to store past frames' features and then distinguish objects with a similar appearance by pixel-level attention-based matching from the memory. 

\subsubsection{Vision Transformer.} Initially proposed for machine translation, Transformers \cite{vaswani2017attention} replace the recurrence and convolutions entirely with hierarchical attention-based mechanisms and achieve outstanding performance. Later, transformer networks became dominant models used in natural language processing (NLP) tasks \cite{wolf2020transformers,zaheer2020big}. Recently, with the observance of its strength in parallel modeling global correlation or attention, transformer blocks were introduced to computer vision tasks, such as image recognition \cite{dosovitskiy2020image}, saliency prediction \cite{zhang2021learning}, object detection \cite{zhu2020deformable,carion2020end}, and object segmentation \cite{wang2021end}, where vision transformers have achieved excellent performance compared to the CNN-based counterparts. Researchers then employed transformer architecture into the VOS task \cite{duke2021sstvos,mao2021joint,TransVOS,AOT}.
SST \cite{duke2021sstvos} adopts the transformer's encoder to compute attention based on the spatial-temporal information among multiple history frames. In \cite{mao2021joint}, a transductive branch is used to capture the spatial-temporal information, which is integrated with an online inductive branch within a unified framework. TransVOS \cite{TransVOS} introduces a transformer-based VOS framework with intuitive structure from the transformer networks in NLP. AOT \cite{AOT} proposes an Identification Embedding to construct multi-object matching and computes attention for multiple objects simultaneously.
In this paper, we introduce a novel bilateral attention transformer framework, where it computes the attention with both the encoded appearance features and the motion features in consideration. Therefore, it is robust to occlusion, drift, and ambiguity between objects with a similar appearance.

\subsubsection{Optical Flow.} Applying optical flow to VOS can encourage motion consistency through the entire video sequence. Early approaches \cite{tsai2016video,xu2018dynamic,SegFlow} consider VOS and optical flow estimation simultaneously with the assumption that the two tasks are complementary.  Recently, RMNet \cite{RMNet} introduces using optical flow generated with an offline model to warp object mask from the previous frame to the query frame and then performing regional matching. It avoids unnecessary matching in regions without target objects or mismatching of objects with a similar appearance. Instead of simply warping the object's mask to indicate the target area, our {\shortname} computes the correlation of each pair of tokens considering their optical flow estimation, appearance similarity, and spatial distance simultaneously. Thus, it is more effective in removing irrelevant matching tokens compared to \cite{RMNet}. Meanwhile, our method is more robust to the accumulated error in warping from the optical flow estimation. % and efficient

\section{Method}
\begin{figure}[t]
\centering
\includegraphics[width=\textwidth, trim={0 1cm 0 0},clip]{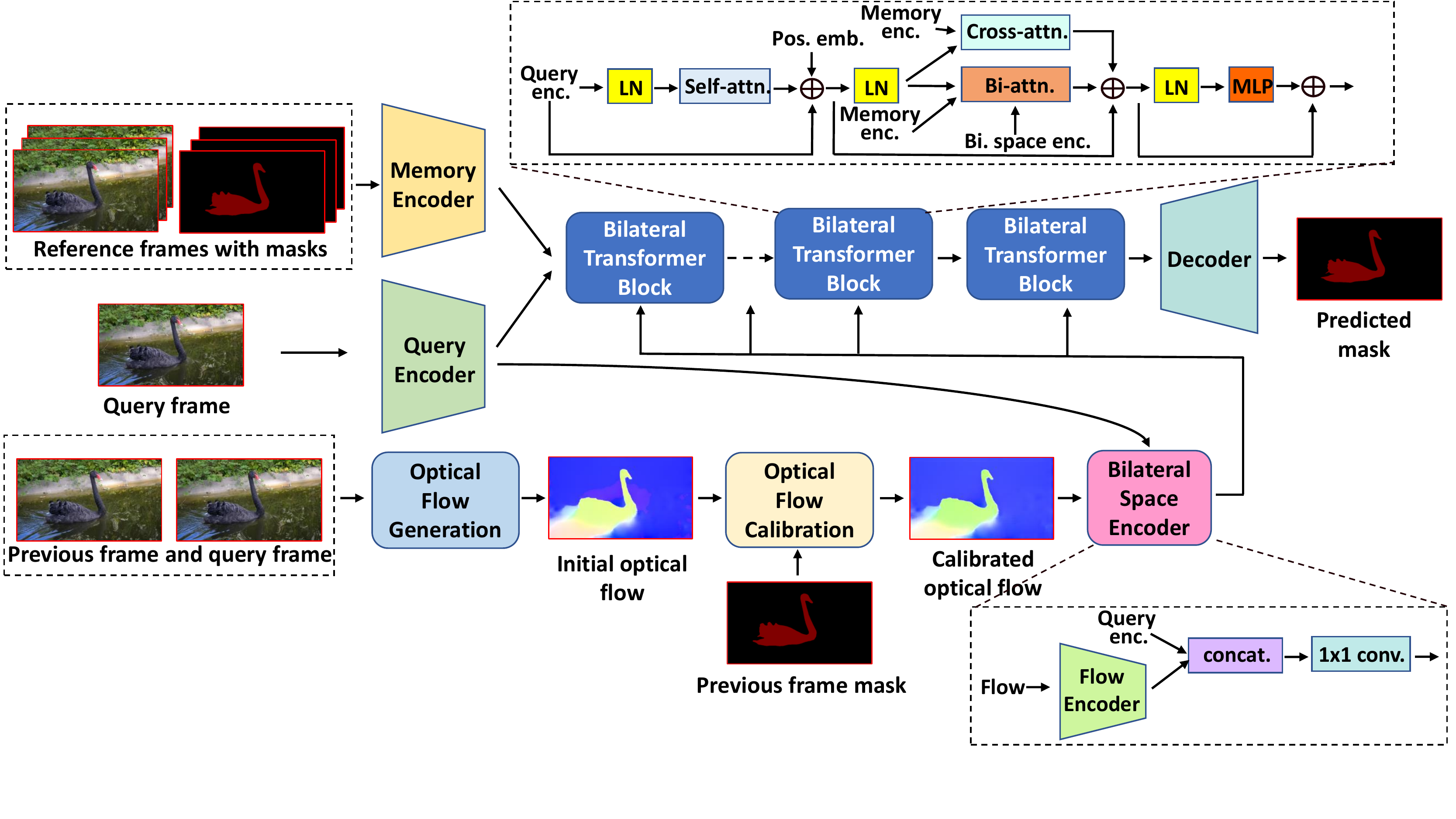}
\caption{Overview of the {\shortname} architecture. Frame-level features of the reference frames and the query frame are extracted through the memory and query encoders, respectively. A pre-trained FlowNet is used to generate an initial optical flow estimation between the previous frame and the query frame, which is then improved by the optical flow calibration module. A bilateral space encoder is used to encode the query features and the calibrated optical flow into a bilateral space encoding, which is used by the bilateral attention. Multiple layers of bilateral transformer blocks are stacked for matching the correspondence between the reference and query features. Lastly, a decoder is used to predict the query frame segmentation mask}
\label{fig:architecture}
\end{figure}
In this section, we first introduce the proposed {\shortname} architecture, and then discuss in depth its core modules: bilateral attention and optical flow calibration.

\subsection{Bilateral Attention Transformer in Motion-Appearance Neighboring space (BATMAN)}
Fig.~\ref{fig:architecture} provides an overview of the proposed {\shortname} architecture. We first extract frame-level features through the memory and query encoders (details in Sec.~\ref{sec:details}) to capture the target object features for establishing correspondence in the later transformer layers. Meanwhile, we compute the initial optical flow between the query frame and its previous frame through a frozen pre-trained FlowNet \cite{teed2020raft}. Then, we feed the object mask from the previous frame, together with the initial optical flow estimation, into our optical flow calibration module to improve the optical flow (Sec.~\ref{sec:calib}). We then encode the calibrated optical flow and the query frame features into tokens in the bilateral space of motion and appearance. Following this, we stack multiple bilateral transformer blocks to model the spatial-temporal relationships among the reference and query frames at pixel-level, based on the bilateral space encoding tokens (Sec.~\ref{sec:bilateral}). After aggregating the spatial-temporal information, the decoder predicts an object mask for the query frame.

\subsection{Bilateral transformer and bilateral attention}\label{sec:bilateral}
As shown in Fig.~\ref{fig:architecture}, in each bilateral transformer block, the query frame features first go through a self-attention \cite{vaswani2017attention} to aggregate the information within the query frame followed by adding a sinusoidal position embedding \cite{vaswani2017attention} encoding the tokens' relative positions. Then we apply cross-attention and bilateral attention (described below) to it with the reference frame features and add the results. Following the common practice in vision transformers \cite{AOT,TransVOS}, we insert layer normalization~\cite{ba2016layer} before and after each attention module. Finally, we employ a two-layer feed-forward MLP block before feeding the output to the next layer.

\subsubsection{Bilateral space encoding} ($E$) is used to index each position (token) of the query frame features in the bilateral space. As shown in Fig.~\ref{fig:architecture}, we first encode the calibrated optical flow using a flow encoder (details in Sec.~\ref{sec:details}). Then we concatenate the optical flow encoding and the query image encoding (from query encoder) in channel dimension. Finally, we use a $1\times1$ convolutional layer to project the concatenation to a 1-dimensional space (in channel) where each position (token) has a single scalar coordinate for the bilateral space of motion and appearance. Bilateral space encoding is employed in bilateral attention below.

\subsubsection{Bilateral attention} is used to aggregate spatial-temporal information between the query tokens and neighboring key tokens from the reference frames in the bilateral space of motion and appearance. Unlike global cross-attention where each query token computes attention with all key tokens from the reference frames, our bilateral attention adaptively selects the most relevant key tokens for each query token based on the bilateral space encoding. To formulate, we define query tokens $Q\in \mathbb{R}^{HW\times C}$, key tokens $K\in \mathbb{R}^{HW\times C}$, and value embedding tokens $V\in \mathbb{R}^{HW\times C}$, where $Q$ is from the query frame and $K$ and $V$ are aggregated from multiple reference frames. $H$, $W$, and $C$ represent the height, width, and channel dimensions of the tokens, respectively. Mathematically, we define bilateral attention as,

\begin{equation} \label{eq:attention}
	BiAttn(Q,K,V) = softmax(\frac{QK^TM}{\sqrt{C}})V
\end{equation}

{\parindent0pt where} $M\in[0,1]^{HW\times HW}$ is the bilateral space binary mask that defines the attention scope for each query token. For each query token $Q_{h,w}$ at $(h,w)$ position, we define the corresponding bilateral space binary mask $M_{h,w}$ as, 

\begin{equation} \label{eq:mask}
	M_{h,w}(i,j,E) = 
	\begin{cases}
      1 & \text{if $|i-h|\leqslant W_d$ and $|j-w|\leqslant W_d$} \\
      & \text{and $|argsort_{W_d}(E_{h,w})-argsort_{W_d}(E_{i,j})| \leqslant W_b$} \\
      0 & \text{otherwise}
    \end{cases}       
\end{equation}

{\parindent0pt where} (i, j) is the position for each key token, $E\in \mathbb{R}^{HW\times 1}$ is the bilateral space encoding of the queries discussed above, $W_d$ and $W_b$ are predefined local windows in spatial and bilateral domains, respectively. $argsort_{W_d}(E_{i,j})$ denotes sorting all bilateral space encoding $E$ within the spatial local window $W_d$ and finding the corresponding index at position $(i,j)$. To train the bilateral space encoding $E$ by stochastic gradient descent directly, in practice, instead of computing $QK^TM$ as shown in Eq.~\ref{eq:attention}, we compute $QK^T+E$ if $M=1$, while computing $QK^T-L$ if $M=0$, where $L\in\mathbb{R}$ is a large positive number. This approximates to $QK^TM$ in Eq.~\ref{eq:attention} after using $softmax$. Eq.~\ref{eq:mask} shows that for each query token, it computes the attention with another key token only if they are close to each other spatially and share similar bilateral space encoding (similar motion and appearance). We further analyze the bilateral space binary mask with visualization in Sec.~\ref{sec:ablation}.

We implement the bilateral attention modules via a multi-headed formulation~\cite{vaswani2017attention} where we linearly project queries, keys, and values multiple times with different learnable projections, and we feedforward the multiple heads of bilateral attention in parallel followed by concatenation and a linear projection. Mathematically, we define the multi-head bilateral attention as,

\begin{equation} \label{eq:head}
	\begin{split}
	    MultiHead(Q,K,V) & = Concat(head_1,...,head_h)W^O\\
	    where~head_i & = BiAttn(QW_i^Q, KW_i^K, VW_i^V)
	\end{split}
\end{equation}

{\parindent0pt where} projection matrices are $W_i^Q\in \mathbb{R}^{C\times d_{hidden}}$, $W_i^K\in \mathbb{R}^{C\times d_{hidden}}$, $W_i^V\in \mathbb{R}^{C\times d_{hidden}}$, and $W^O\in \mathbb{R}^{C\times C}$. In this work, we set the number of heads ($h=C/d_{hidden}$) to 8~\cite{vaswani2017attention}, where $d_{hidden}$ is the hidden dimension of each head.

\subsection{Optical flow calibration}\label{sec:calib}
\begin{figure}[tp]
    \centering
    \includegraphics[width=\textwidth, trim={2.3cm 1.5cm 3cm 3cm},clip]{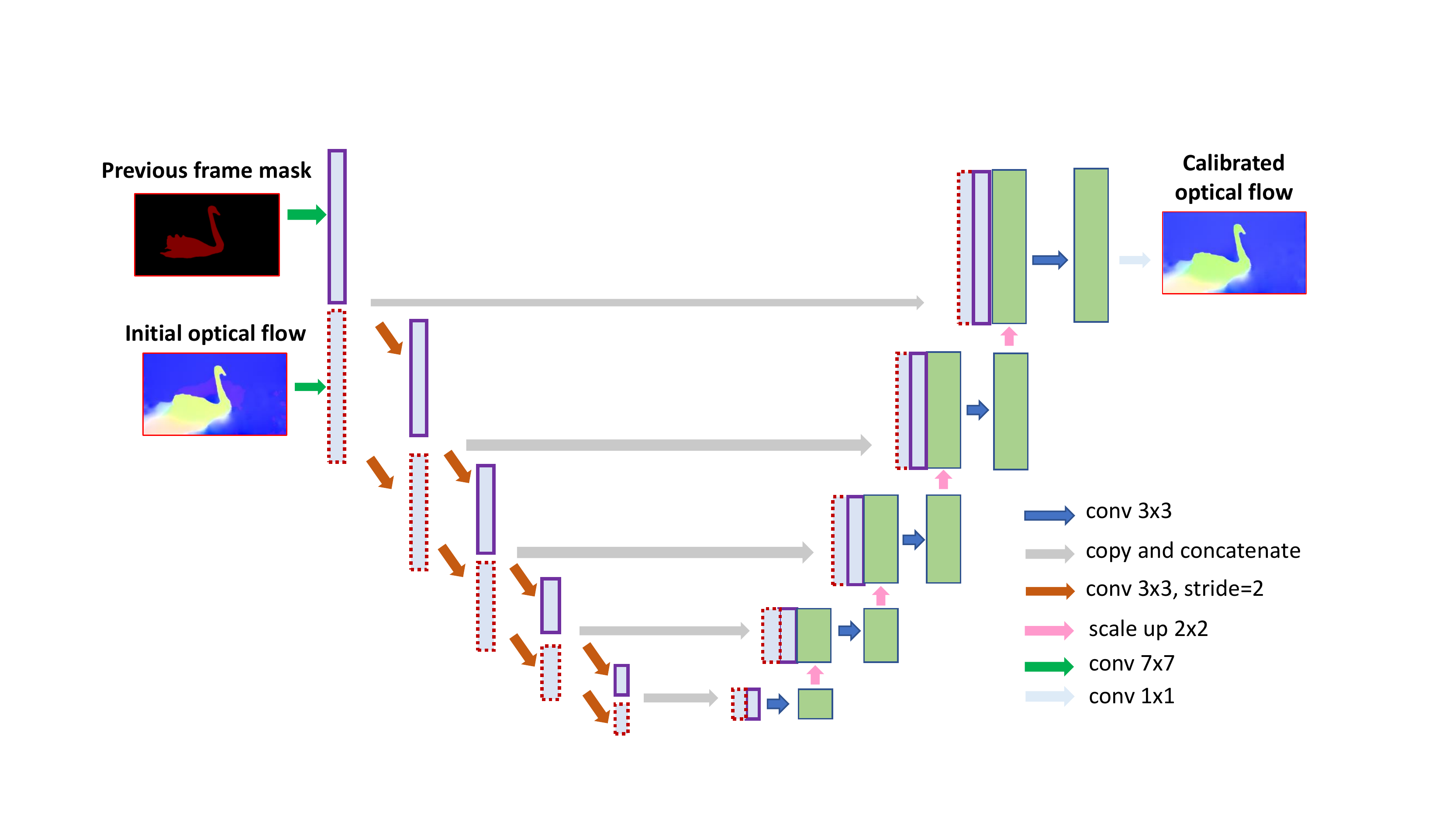}
    \caption{The optical flow calibration module. A CNN in the U-Net architecture~\cite{ronneberger2015u} is used to fuse the segmentation mask into the optical flow}
    \label{fig:flowCalib}
\end{figure}

As mentioned in the introduction, optical flow estimation can be noisy for objects with large motion and in texture-less areas. We introduce an optical flow calibration module to improve flow estimation by leveraging the segmentation mask from the previous frame. As shown in Fig.~\ref{fig:flowCalib}, the module employs an architecture similar to U-Net~\cite{ronneberger2015u} with 11-layers total.  
To train this module to improve optical flow, we compute the Mean Square Error (MSE)
between the initial optical flow and the output optical flow in training.
Without the MSE loss,
mask information can dominate the calibration module and thereby generate an embedding feature for the mask instead. 

\section{Experiments}

\setlength{\tabcolsep}{4pt}
\begin{table}[t]
\begin{center}
\caption{Results on Youtube-VOS 2019/2018 validation split. Subscript $s$ and $u$ denote scores in seen and unseen categories, respectively. {\shortname} outperforms all state-of-the-art methods on both benchmarks}
\label{table:youtubevos}
\begin{tabular}{cccccc|ccccc}
\hline
\multirow{2}{4em}{Method} & \multicolumn{5}{c}{Youtube-VOS 2019} & \multicolumn{5}{c}{Youtube-VOS 2018} \\
& $\mathcal{J}\&\mathcal{F}$ & $\mathcal{J}_s$ & $\mathcal{J}_u$ & $\mathcal{F}_s$ & $\mathcal{F}_u$ & $\mathcal{J}\&\mathcal{F}$ & $\mathcal{J}_s$ & $\mathcal{J}_u$ & $\mathcal{F}_s$ & $\mathcal{F}_u$\\
\hline
STM\cite{STM} & - & - & - & - & - & 79.4 & 79.7 & 72.8 & 84.2 & 80.9\\
AFB-URR\cite{liang2020video} & - & - & - & - & - & 79.6 & 78.8 & 74.1 & 83.1 & 82.6\\
KMN\cite{seong2020kernelized} & - & - & - & - & - & 81.4 & 81.4 & 75.3 & 85.6 & 83.3\\
CFBI\cite{yang2020collaborative} & 81.0 & 80.6 & 75.2 & 85.1 & 83.0 & 81.4 & 81.1 & 75.3 & 85.8 & 83.4\\
LWL\cite{bhat2020learning} & - & - & - & - & - & 81.5 & 80.4 & 76.4 & 84.9 & 84.4\\
RMN\cite{RMNet} & - & - & - & - & - & 81.5 & 82.1 & 75.7 & 85.7 & 82.4\\
SST\cite{duke2021sstvos} & 81.8 & 80.9 & 76.6 & - & - & 81.7 & 81.2 & 76.0 & - & -\\
TransVOS\cite{TransVOS} & - & - & - & - & - & 81.8 & 82.0 & 75.0 & 86.7 & 83.4\\
LCM\cite{hu2021learning} & - & - & - & - & - & 82.0 & 82.2 & 75.7 & 86.7 & 83.4 \\
CFBI+\cite{yang2021collaborative} & 82.6 & 81.7 & 77.1 & 86.2 & 85.2 & 82.8 & 81.8 & 77.1 & 86.6 & 85.6\\
STCN\cite{STCN} & 82.7 & 81.1 & 78.2 & 85.4 & 85.9 & 83.0 & 81.9 & 77.9 & 86.5 & 85.7\\
RPCMVOS\cite{RPCMVOS} & 83.9 & 82.6 & \textbf{79.1} & 86.9 & 87.1 & 84.0 & 83.1 & 78.5 & 87.7 & 86.7\\
AOT\cite{AOT} & 84.1 & 83.5 & 78.4 & 88.1 & 86.3 & 84.1 & 83.7 & 78.1 & 88.5 & 86.1\\
\hline
\textbf{\shortname} & \textbf{85.0} & \textbf{84.5} & 79.0 & \textbf{89.3} & \textbf{87.2} & \textbf{85.3} & \textbf{84.7} & \textbf{79.2} & \textbf{89.8} & \textbf{87.4}\\
\hline
\end{tabular}
\end{center}
\end{table}

We validate {\shortname} on popular benchmark datasets YouTube-VOS 2019/2018 and DAVIS 2017/2016. We first provide implementation details, followed by the experimental results. We then present the ablation study on our design.

\subsection{Implementation details}\label{sec:details}
We use ResNet50 \cite{he2016deep} as the feature extractor for memory/query/flow encoder. We follow the identification embedding in \cite{AOT} to encode multiple object masks in the memory encoding simultaneously. We use a RAFT \cite{teed2020raft} model pre-trained on FlyingThings3D \cite{MIFDB16} for optical flow generation. We use FPN \cite{lin2017feature} with Group Normalization \cite{wu2018group} as the decoder. We employ $12$ bilateral transformer blocks with $W_d$ and $W_b$ set to $7$~\cite{AOT} and $84$ (details in supplementary), respectively.

We implement our model in PyTorch \cite{paszke2017automatic} and train with a batch size of 16 distributed on 8 V100 GPUs. Following previous works \cite{lu2020video,AOT,TransVOS,RMNet}, we first pre-train our model on synthetic video sequences generated from static image datasets (COCO \cite{lin2014microsoft}, ECSSD \cite{shi2015hierarchical}, MSRA10K \cite{cheng2014global}, SBD \cite{hariharan2011semantic}, PASCALVOC2012 \cite{everingham2010pascal}) by applying random augmentations. We then train the model on the VOS benchmarks. 
The loss function is a combination of bootstrapped cross-entropy loss, soft Jaccard loss \cite{6909471}, and mean squared error loss. 
The training is optimized using AdamW \cite{loshchilov2017decoupled} optimizer and Exponential Moving Average (EMA) \cite{polyak1992acceleration}. The learning rate for training is set to $2\times 10^{-4}$ with a weight decay of $0.07$. We train the model for 100,000 iterations.

\subsection{Experimental results}
We present validation results on the popular Youtube 2019/2018 and DAVIS 2017/2016 benchmarks compared to existing state-of-the-art methods.

\setlength{\tabcolsep}{4.2pt}
\begin{table}[h]
\begin{center}
\caption{Comparisons to the state-of-the-art methods on DAVIS benchmarks. (\textbf{Y}) indicates including Youtube-VOS dataset in training. {\shortname} outperforms all state-of-the-art methods on all three DAVIS benchmarks}
\label{table:davis}
\begin{tabular}{cccc|ccc|ccc}
\hline
\multirow{2}{3.8em}{Method} & \multicolumn{3}{c}{DAVIS 2017 val} & \multicolumn{3}{c}{DAVIS 2017 test-dev} & \multicolumn{3}{c}{DAVIS 2016 val} \\
& $\mathcal{J}\&\mathcal{F}$ & $\mathcal{J}$ & $\mathcal{F}$ & $\mathcal{J}\&\mathcal{F}$ & $\mathcal{J}$ & $\mathcal{F}$ & $\mathcal{J}\&\mathcal{F}$ & $\mathcal{J}$ & $\mathcal{F}$\\
\hline
AFB-URR\cite{liang2020video} & 74.6 & 73.0 & 76.1 & - & - & - & - & - & -\\
LWL\cite{bhat2020learning} & 81.6 & 79.1 & 84.1 & - & - & - & - & - & -  \\
STM\cite{STM}(\textbf{Y}) & - & 79.2 & 84.3 & - & - & - & - & 88.7 & 89.9 \\
CFBI\cite{yang2020collaborative}(\textbf{Y}) & 81.9 & 79.3 & 84.5 & 75.0 & 71.4 & 78.7 & 89.4 & 88.3 & 90.5  \\
SST\cite{duke2021sstvos}(\textbf{Y}) & 82.5 & 79.9 & 85.1 & - & - & - & - & - & -  \\
KMN\cite{seong2020kernelized}(\textbf{Y}) & 82.8 & 80.0 & 85.6 & 77.2 & 74.1 & 80.3 & 90.5 & 89.5 & 91.5 \\
CFBI+\cite{yang2021collaborative}(\textbf{Y}) & 82.9 & 80.1 & 85.7 & 75.6 & 71.6 & 79.6 & 89.9 & 88.7 & 91.1  \\
RMN\cite{RMNet}(\textbf{Y}) & 83.5 & 81.0 & 86.0 & 75.0 & 71.9 & 78.1 & 88.8 & 88.9 & 88.7  \\
LCM\cite{hu2021learning}(\textbf{Y}) & 83.5 & 80.5 & 86.5 & 78.1 & 74.4 & 81.8 & 90.7 & 89.9 & 91.4\\
RPCMVOS\cite{RPCMVOS}(\textbf{Y}) & 83.7 & 81.3 & 86.0 & 79.2 & 75.8 & 82.6 & 90.6 & 87.1 & 94.0  \\
TransVOS\cite{TransVOS}(\textbf{Y}) & 83.9 & 81.4 & 86.4 & 76.9 & 73.0 & 80.9 & 90.5 & 89.8 & 91.2  \\
AOT\cite{AOT}(\textbf{Y}) & 84.9 & 82.3 & 87.5 & 79.6 & 75.9 & 83.3 & 91.1 & 90.1 & 92.1  \\
STCN\cite{STCN}(\textbf{Y}) & 85.4 & 82.2 & 88.6 & 76.1 & 72.7 & 79.6 & 91.6 & \textbf{90.8} & 92.5 \\
\hline
\textbf{\shortname}(\textbf{Y}) & \textbf{86.2} & \textbf{83.2} & \textbf{89.3} & \textbf{82.2} & \textbf{78.4} & \textbf{86.1} & \textbf{92.5} & 90.7 & \textbf{94.2} \\
\hline
\end{tabular}
\end{center}
\end{table}

\noindent\subsubsection{Metrics.}
The region similarity ($\mathcal{J}$) and the boundary accuracy ($\mathcal{F}$) are computed following the standard evaluation setting proposed in~\cite{perazzi2016benchmark}. On DAVIS, we report the two metrics and their mean value ($\mathcal{J\&F}$). On YouTube-VOS, we report all the metrics on seen categories and unseen categories separately % and report them.
as generated by the evaluation server at CodaLab.

\noindent\subsubsection{Youtube-VOS~\cite{xu2018youtube}} is a large-scale dataset for multi-object video segmentation with objects in multiple categories. In YouTube-VOS 2018, the \textsl{training} set contains $3,471$ videos with $5,945$ unique objects in 65 categories and the \textsl{validation} set has $474$ videos containing of $894$ unique objects in $65$ seen categories and additional $26$ unseen categories. YouTube-VOS 2019 expands the YouTube-VOS 2018 dataset with more videos and object annotations. Its \textsl{training} set contains the same $3,471$ videos but has $6,459$ objects. Its \textsl{validation} set has $507$ videos containing of $1,063$ objects. With the existence of the unseen object categories, the YouTube-VOS is useful to evaluate the generalization capability of the VOS model on unseen object categories. We evaluate all the results on the official YouTube-VOS evaluation servers on CodaLab.

Table~\ref{table:youtubevos} shows that {\shortname} outperforms all state-of-the-art on Youtube-VOS 2019 and 2018 benchmarks. The higher region similarity ($\mathcal{J}$) and better boundary accuracy ($\mathcal{F}$) validate that bilateral attention is able to learn the most informative features from the reference frames and match the query frames.

\begin{figure}[t]
\centering
\includegraphics[width=\textwidth, trim={1.5cm 1.5cm 2cm 1cm},clip]{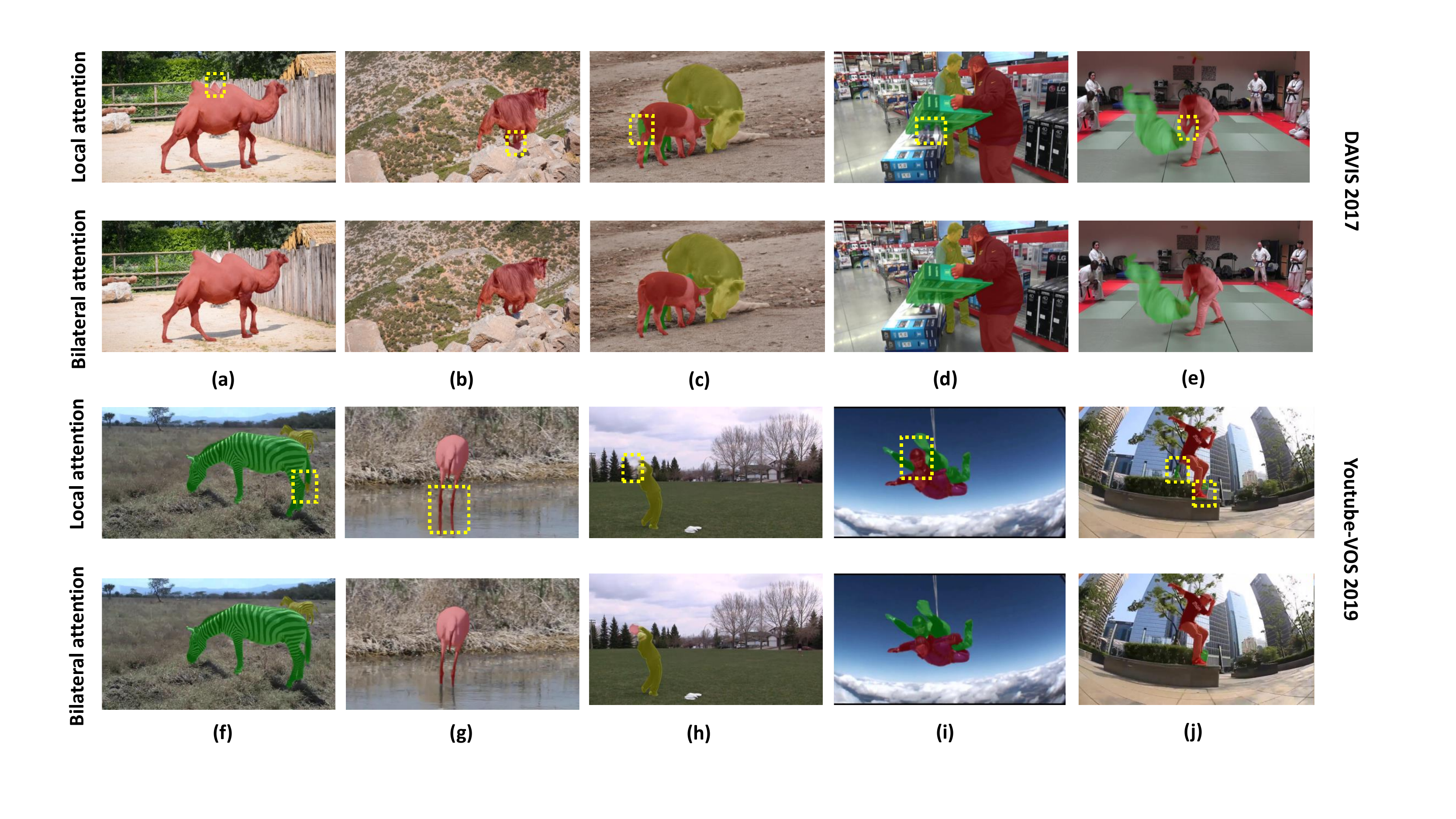}
%\vspace{-1em}
\caption{Qualitative results. Compared to spatial local attention, bilateral attention segments objects better especially when background shares similar appearance with the target object}
\label{fig:segmentation}
%\vspace{-0.2cm}
\end{figure}

\subsubsection{DAVIS} is one of the most popular benchmarks for video object segmentation with high-quality masks for salient objects. As part of DAVIS, DAVIS 2016 \cite{perazzi2016benchmark} is a single-object benchmark and DAVIS 2017 \cite{pont20172017} is a multi-object extension of DAVIS 2016. In DAVIS 2016, the \textsl{training} and \textsl{validation} sets contain 30 and 20 videos, respectively. In DAVIS 2017, the \textsl{training} set consists of 60 videos, and the \textsl{validation} set consists of 30 videos, and the \textsl{test-dev} set consists of 30 videos with only the first frame annotated. 

Table~\ref{table:davis} compares {\shortname} with existing state-of-the-art methods on DAVIS 2017 validation set, test-dev set, and DAVIS 2016 validation set. Note that KMN \cite{seong2020kernelized} only reports the results of DAVIS 2017 test-dev split with images resized to 600p. We follow the standard practice of most previous works and keep the images in the original 480p resolution in evaluation. On both multi-object datasets (DAVIS2017 val/test) and single-object dataset (DAVIS 2016), {\shortname} outperforms all existing state-of-the-art methods. Moreover, {\shortname} achieves the largest absolute accuracy improvement (2.6\%) on the hardest DAVIS 2017 test-dev split, which validates the robustness of our model for VOS.

\subsection{Ablation study} \label{sec:ablation}
In this section, we analyze the effectiveness of the bilateral attention and compare it to the conventional spatial local attention, as well as the efficacy of the calibrated optical flow. For qualitative analysis, we visualize the bilateral space binary mask generated by the bilateral attention, and the optical flow output from our calibration module.

\setlength{\tabcolsep}{6pt}
\begin{table}[t]
% \begin{center}
\centering
\caption{Ablation on bilateral attention. The model with bilateral attention outperforms that with spatial local attention on all benchmarks}
\label{table:ablation}
\begin{tabular}{cccccc}
\hline
\multirow{2}{*}{\makecell[c]{Attention\\type}} & \multirow{2}{*}{\makecell[c]{DAVIS\\2017 val}} & \multirow{2}{*}{\makecell[c]{DAVIS 2017\\test-dev}} & \multirow{2}{*}{\makecell[c]{DAVIS\\2016 val}} & \multirow{2}{*}{\makecell[c]{Youtube-\\VOS 2019}} & \multirow{2}{*}{\makecell[c]{Youtube-\\VOS 2018}} \\
\\
\hline
Spatial local & 84.9 & 77.5 & 91.6 & 84.1 & 83.8 \\
Bilateral & \textbf{86.2} & \textbf{82.2} & \textbf{92.5} & \textbf{85.0} & \textbf{85.3}\\
\hline
\end{tabular}
% \end{center}
\end{table}

\begin{figure}[t]
\centering
\includegraphics[width=\textwidth]{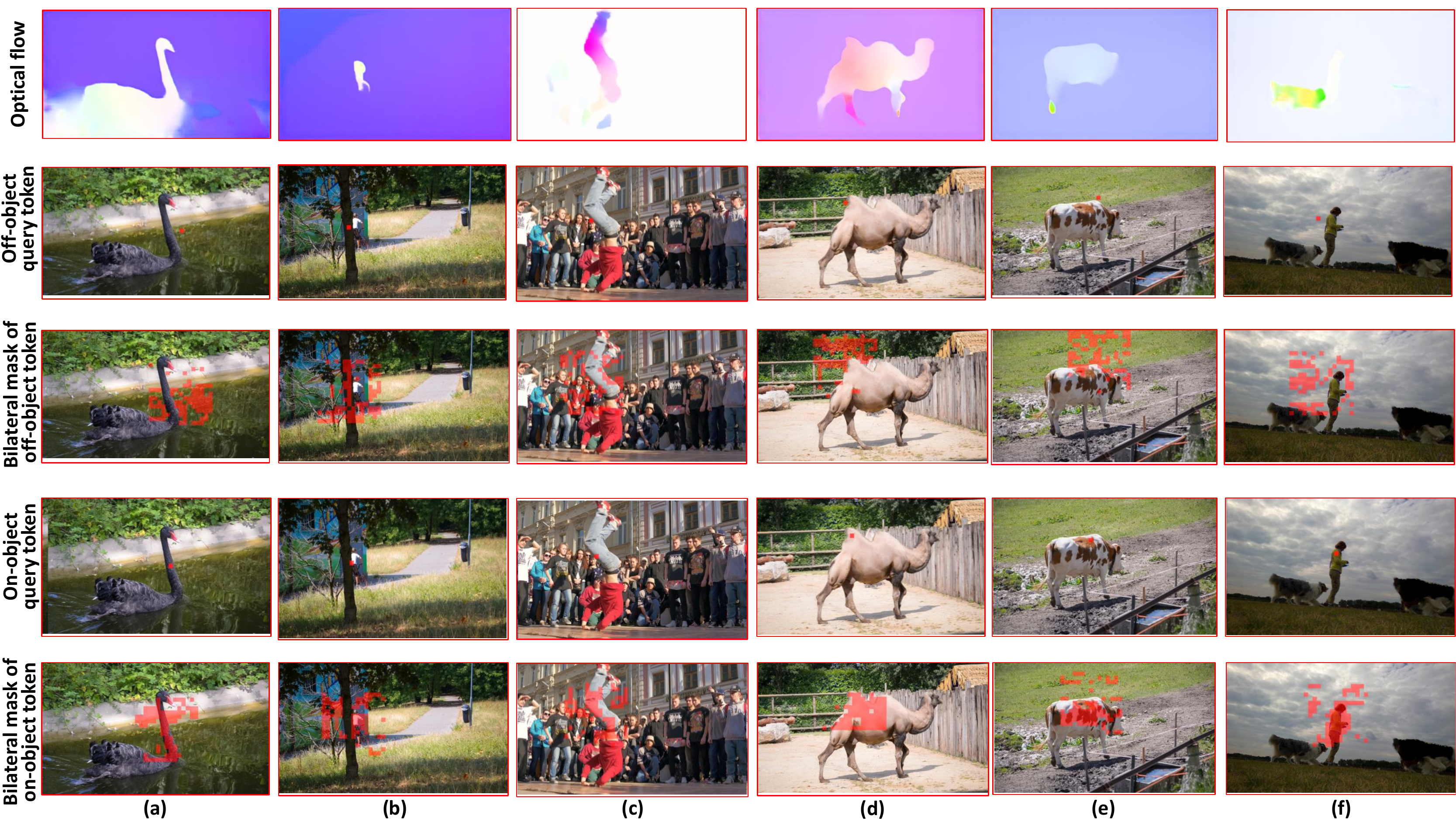}
%\vspace{-1em}
\caption{Visualization of bilateral space binary masks from the bilateral attention. The bilateral attention adaptively generates binary masks for on and off object query tokens. Better view in color version}
\label{fig:mask}
%\vspace{-1em}
\end{figure}

\begin{figure}[t]
    \centering
    \begin{subfigure}{\textwidth}
        \centering
        \includegraphics[width=0.98\textwidth]{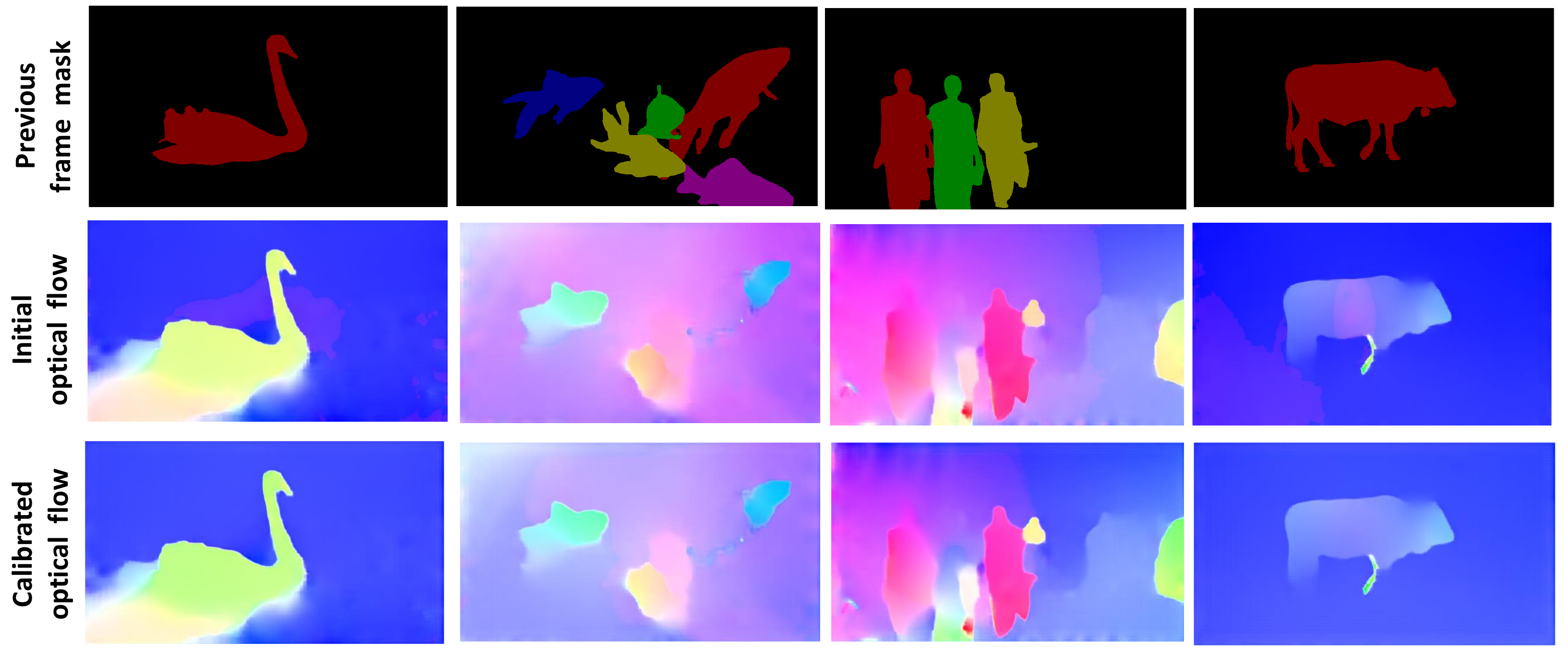}
        \caption{Comparison of the initial optical flow (middle) and the calibrated optical flow (bottom). The calibrated optical flow is smoother within the same object, and sharper at object boundary}
    \end{subfigure}
    %\vspace{1em}
    \begin{subfigure}{\textwidth}
        \centering
        \includegraphics[width=0.58\textwidth]{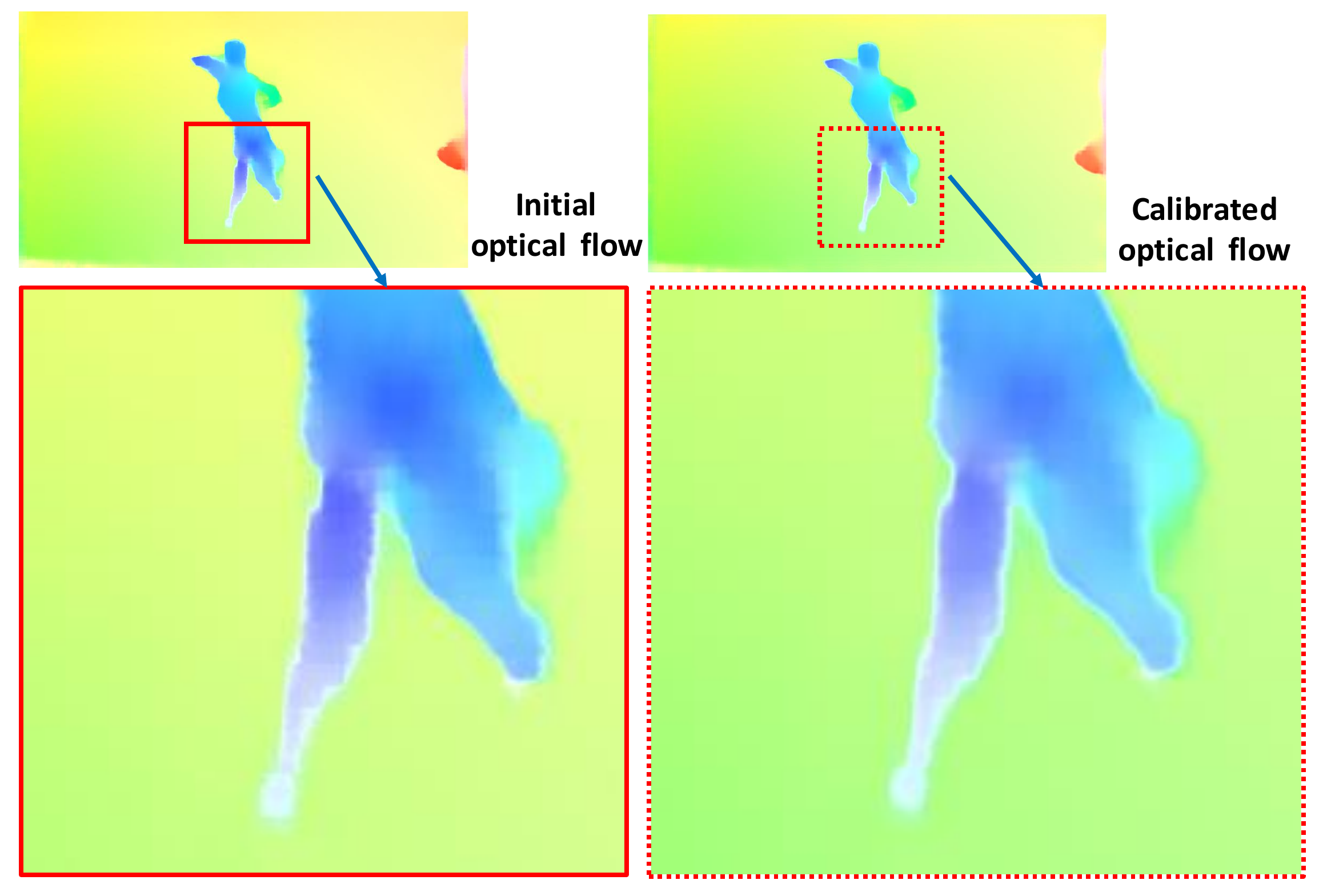}
        \caption{Blocky artifact on the initial optical flow is decreased on the calibrated optical flow}
    \end{subfigure}
    %\vspace{-1.5em}
    \caption{Visualization of optical flow on Davis 2017 \textsl{val.} set. The calibrated optical flow is smoother within the object and sharper at the boundary}
    \label{fig:flowCalib-example}
\end{figure}

\subsubsection{Bilateral Attention.} 
Table~\ref{table:ablation} compares the accuracy ($\mathcal{J}\&\mathcal{F}$) between our proposed bilateral attention and the conventional spatial local attention, and validates that the bilateral attention achieves superior performance on all benchmarks. We also visualize the segmentation masks from the two attention mechanisms in Fig.~\ref{fig:segmentation} for both DAVIS 2017 and Youtube-VOS 2019. We can see that with spatial local attention, the model tends to fail to segment objects with similar appearances~(e.g., the second camel is included in the mask of the first camel (Fig.~\ref{fig:segmentation}a); part of the red pig is segmented as the green pig (Fig.~\ref{fig:segmentation}c); the right hand of the man in green is mistakenly segmented as part of the man in red (Fig.~\ref{fig:segmentation}e); the tail of the zebra in the green mask is mistakenly segmented as that of the zebra in yellow (Fig.~\ref{fig:segmentation}f)). Besides, when the appearance features (especially at the object boundary) are fuzzy~(e.g., the shade of the goat (Fig.~\ref{fig:segmentation}b), the reflection on the TV box (Fig.~\ref{fig:segmentation}d), and the reflection of the bird's legs in the water (Fig.~\ref{fig:segmentation}g)), the model with spatial local attention finds it difficult to segment the object properly. 
In contrast, the bilateral attention and the resultant adaptive bilateral space binary masks enables our model to segment target objects correctly, especially when the target object exhibits salient motion (e.g., the Frisbee (Fig.~\ref{fig:segmentation}h) and the skydiving men (Fig.~\ref{fig:segmentation}i)). We provide additional visualizations for segmentation in the supplementary.

Fig.~\ref{fig:mask} shows some examples of the binary masks generated from the bilateral attention. The first row shows the optical flow of the query frames. One off-object (background) query token is highlighted (in red) in the second row for each scene. The corresponding bilateral space binary mask is highlighted in the third row. In comparison, we also show an on-object query token in the fourth row, and the corresponding binary mask is given in the last row. We can see that for an off-object query token, the bilateral attention module tends to focus on the background locations (e.g., the water around the swan neck (Fig.~\ref{fig:mask}a) or the sky around the woman with dogs (Fig.~\ref{fig:mask}f)). On the other hand, when the query token is on the object, it tends to select the neighboring on-object tokens (e.g., the leg of the dancing man (Fig.~\ref{fig:mask}c) or the camel hump (Fig.~\ref{fig:mask}d)) for the attention computation. This qualitatively validates that adaptive attention computation enables propagating segmentation masks from the reference frames to the query frame more accurately.

\subsubsection{Optical flow calibration.}
The optical flow calibration module leverages the predicted previous frame mask %$M^{t-1}$ 
to improve the optical flow estimation for the current frame. Table~\ref{table:calib} compares the bilateral attention w/ and w/o calibrated optical flow. With the calibrated optical flow, {\shortname} achieves higher accuracy on all benchmarks, validating that optical flow is improved with the help of the previous frame segmentation mask. As shown in Fig. \ref{fig:flowCalib-example}, the calibrated optical flow is smoother, both within the same object and within the background. Meanwhile, the object boundary is sharper. Specifically, the blocky artifacts along the object boundary, which exists in the initial optical flow, are reduced effectively without affecting the object boundary sharpness.  

\setlength{\tabcolsep}{6pt}
\begin{table}[ht]
\begin{center}
\caption{Comparisons of bilateral attention w/ and w/o optical flow calibration. Calibrating the optical flow leads to higher accuracy on all benchmarks}
\label{table:calib}
\begin{tabular}{cccccc}
\hline
\multirow{2}{*}{\makecell[c]{Optical\\flow type}} & \multirow{2}{*}{\makecell[c]{DAVIS\\2017 val}} & \multirow{2}{*}{\makecell[c]{DAVIS 2017\\test-dev}} & \multirow{2}{*}{\makecell[c]{DAVIS\\2016 val}} & \multirow{2}{*}{\makecell[c]{Youtube-\\VOS 2019}} & \multirow{2}{*}{\makecell[c]{Youtube-\\VOS 2018}} \\
\\
\hline
w/o calibration & 86.0 & 81.7 & 92.4 & 84.6 & 84.8 \\
w/ calibration & \textbf{86.2} & \textbf{82.2} & \textbf{92.5} & \textbf{85.0} & \textbf{85.3}\\
\hline
\end{tabular}
\end{center}
\end{table}

\begin{figure}[t]
\centering
\includegraphics[width=\textwidth, trim={2cm 6cm 3cm 0.5cm},clip]{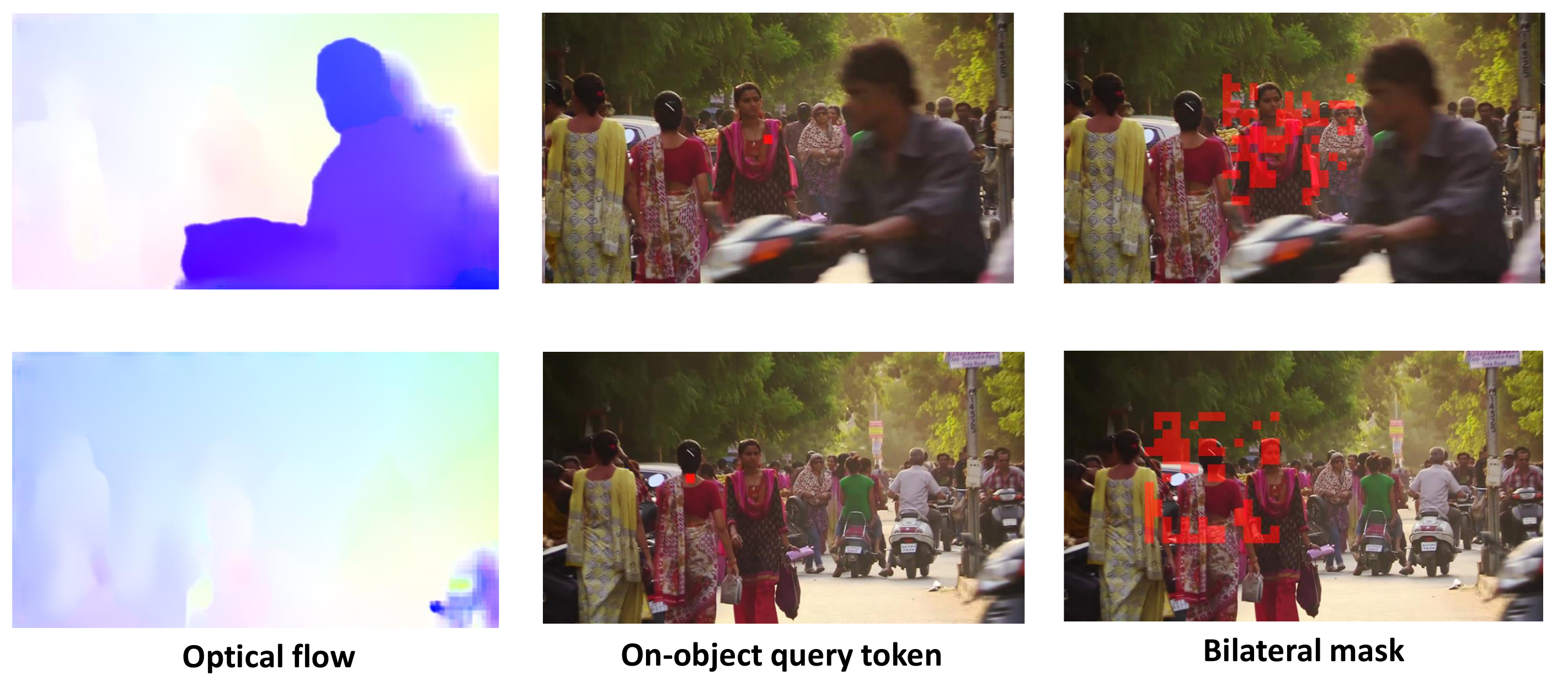}
% \vspace{-1em}
\caption{Failure cases of the bilateral binary mask generation. The bilateral attention may lose focus when a background object exhibits dominant motion and/or the target object does not exhibit salient motion}
\label{fig:limitation}
\end{figure}

\subsubsection{Limitations.}
The bilateral space binary mask generation is influenced by the motion in the scene. Therefore, if the target objects do not exhibit salient motion, or some background object(s) exhibit salient motion and/or share(s) a similar appearance to the target object(s), the bilateral mask can be noisy and the bilateral attention may lose focus. Fig.~\ref{fig:limitation} shows two failure cases: in the upper row, a man on a motorcycle moves quickly across the scene, which overwhelms the motion of the target woman. Hence, the bilateral attention fails to focus on the target object. Similarly, in the bottom row, the motion of the target woman is not salient (especially on the boundary) so the bilateral mask scatters. We plan to extend our method to better handle such scenarios.

\section{Conclusions}
This paper proposes a novel architecture, {\shortname}, for semi-supervised VOS by adaptively computing attention between the query frame and reference frames based on the bilateral encoding of motion and appearance. Compared to conventional spatial local attention, bilateral attention adaptively selects the most relevant tokens to compute the correlation attention which helps to match the object correspondence spatially and temporally with the help of calibrated optical flow. Extensive experiments validate that {\shortname} outperforms all existing state-of-the-art on all popular Youtube-VOS and DAVIS benchmarks.

\clearpage
% ---- Bibliography ----
%
% BibTeX users should specify bibliography style 'splncs04'.
% References will then be sorted and formatted in the correct style.
%
\bibliographystyle{splncs04}
%\bibliography{main}

\end{document}